\documentclass{llncs}

\usepackage{graphicx}
\usepackage{amsmath}
\usepackage{amsfonts}
\usepackage{amssymb}
\usepackage{bm}
\usepackage{mathtools}
\usepackage{nicefrac}
\usepackage{widetable}
\usepackage{booktabs}
\usepackage{minitoc} 
\usepackage[toc,page]{appendix} 
\usepackage{hyperref} 
\usepackage{soul}
\usepackage{esvect}
\usepackage{pgfgantt}
\usepackage{hyperref}
\usepackage{makeidx}
\usepackage[misc]{ifsym}
\usepackage{tikz} 
	\usetikzlibrary{bayesnet} 
    \usetikzlibrary{arrows.meta}
\usepackage{multirow}
\usepackage{caption}
\newcommand*{\cequal}{\overset{c}{=}}

\begin{document}

\frontmatter

\pagestyle{headings}

\mainmatter

\title{Flexible Bayesian Modelling\\for Nonlinear Image Registration}

\author{Mikael Brudfors\inst{1,*} (\Letter) \and Ya\"{e}l 
Balbastre\inst{1,*}
\and Guillaume Flandin\inst{1} \and Parashkev Nachev\inst{2} \and John
Ashburner\inst{1}}
\institute{Wellcome Centre for Human Neuroimaging, UCL, UK\\
\email{mikael.brudfors.15@ucl.ac.uk} \and UCL Institute of Neurology, 
UK \\\vspace{3mm}
\textasteriskcentered\ These authors contributed equally to this work}

\maketitle

\begin{abstract}

We describe a diffeomorphic registration algorithm that allows groups of
images to be accurately aligned to a common space, which we intend to incorporate into the SPM software. The idea is to
perform inference in a probabilistic graphical model that accounts for
variability in both shape and appearance. The resulting framework is
general and entirely unsupervised. The model is evaluated at
inter-subject registration of 3D human brain scans. Here, the main
modeling assumption is that individual anatomies can be generated by
deforming a latent `average' brain. The method is agnostic to imaging
modality and can be applied with no prior processing. We evaluate the
algorithm using freely available, manually labelled datasets. In this
validation we achieve state-of-the-art results, within reasonable 
runtimes, against previous state-of-the-art widely used, inter-subject 
registration algorithms. On the unprocessed dataset, the increase 
in overlap score is over 17\%. These results demonstrate the benefits 
of using informative computational anatomy frameworks for nonlinear 
registration.

\end{abstract}

\section{Introduction}

\begin{figure*}[t]
	\centering
	\includegraphics[width=\textwidth]{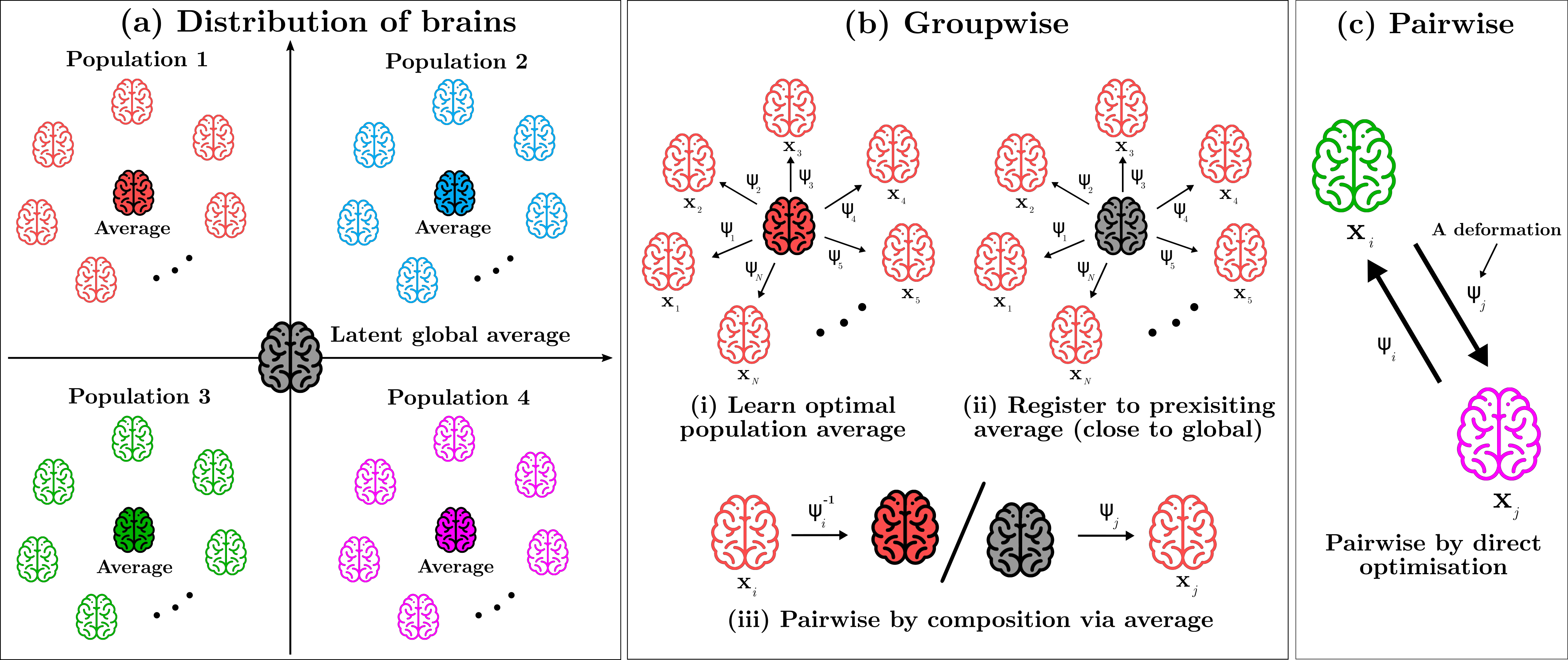}
	\caption{\textbf{(a)} Multiple populations of brain scans have their 
		individual averages; the assumption in this paper is that there exists 
		a latent global average. \textbf{(b)} A groupwise method can either 
		learn the optimal population-specific average (i), or use an already 
		learned average (ii), the closer the learned average is to the global, 
		the better the method should generalise to unseen test data. In both 
		cases, all population scans are deformed towards the average. Pairwise 
		deformations are then obtained by composing deformations via this 
		average (iii).  The proposed approach belongs here, and can be used for 
		both (i) and (ii). \textbf{(c)} A pairwise method directly deforms one 
		image towards another, usually by optimising some similarity metric or by applying a learned function. 
		The common space then consists of just the two images to be registered.
	}%
	\label{fig:RegMethods}
\end{figure*}

This paper presents a flexible framework for registration of 
a population of images into a common space, a procedure known as spatial 
normalisation \cite{friston1995spatial}, or congealing \cite{zollei2005efficient}.  Depending on the quality of the common space, accurate pairwise alignments can be produced by composing deformations that map two subjects to 
this space. The method is 
defined by a joint probability distribution that 
describes how the observed data can be generated. This \emph{generative model}
accounts for both \emph{shape} and \emph{appearance} variability; its conditional dependences 
producing a more robust procedure. 
Shape is encoded by a tissue \textit{template}, that is deformed
towards each image by a subject-specific 
composition of a rigid and a diffeomorphic transform. Performing 
registration on the tissue level, rather than intensity, has been shown 
to be a more robust method of registering medical images 
\cite{heckemann2010improving}. Appearance is encoded by subject-specific
Gaussian mixture models, with prior 
hyper-parameters shared across the population. A key assumption of the model is that there exists a latent average representation (\emph{e.g.}, brain), this is illustrated in Fig. \ref{fig:RegMethods}a. 

Images of human organs differ in their morphology, the goal of 
spatial normalisation is to deform individual organs so that anatomical locations 
correspond between different subjects (a selective removal of the inter-individual anatomical variance). The 
deformations that are computed from this inter-subject registration 
therefore capture meaningful individual shape information. Although 
\emph{not} constrained to a specific organ, our method will here be applied 
to spatially normalise brain magnetic resonance images (MRIs). Spatial 
normalisation is a critical first step in many neuroimaging analyses, \emph{e.g.}, the comparison of tissue composition \cite{draganski2004changes}
or functional MRI activation 
\cite{fox1995spatial} across individuals; shape mapping 
\cite{csernansky1998hippocampal}; the extraction of predictive features for 
machine learning tasks \cite{mourao2012individualized}; or the identification 
of lesions 
\cite{seghier2008lesion}. The success of these tasks is therefore 
fundamentally coupled with the quality of the inter-individual alignment. 
Neuroimaging meta-analysis 
\cite{yarkoni2011large} is another research area that relies on spatial 
normalisation. Currently, statistical maps are coarsely registered into the 
MNI space. Better normalisation towards a more generic, multi-modal, 
high-resolution space could greatly improve the power and spatial specificity 
of such meta-analyses. 

In general, registration tasks can be 
classified as either pairwise or groupwise (Fig. \ref{fig:RegMethods}b-c). Pairwise methods optimise a 
mapping between two images, and only their two spaces exist. Groupwise methods 
aim to align several images into an optimal common space. 
Spatial normalisation aims to register a group of 
images into a pre-existing common space, defined by some average.
Most nonlinear registration methods optimise an energy that
comprises two terms: one that measures the similarity between a deformed
and a fixed image and one that enforces the smoothness of the 
deformation. Two main families emerge, whether they penalise the 
displacement fields (inspired by solid physics) or 
their infinitesimal rate of change (inspired by fluid physics), 
allowing for large diffeomorphic deformations 
\cite{christensen1997volumetric,avants2008symmetric}. 
Concerning the optimisation scheme, a common strategy is to work with energies 
that allow for a probabilistic interpretation 
\cite{ashburner2005unified,ashburner2007fast,andersson2007non,bhatia2007groupwise,vercauteren2009diffeomorphic}. The optimisation can in this case be cast as an inference problem, which is the approach taken in this paper.
More recently, it has been proposed to use deep neural networks 
to learn the normalisation function 
\cite{balakrishnan2019voxelmorph,dalca2019learning,fan2019birnet,krebs2019learning}.
At training time, however, these approaches still use a two-term loss function 
that enforces data consistency while penalising non-smoothness of 
the deformations. These models have demonstrated remarkable 
speed-ups in runtime for volumetric image registration, with similar 
accuracies to the more classical methods. 

Note that all of the above methods either require some sort of prior image processing or are 
restricted to a specific MR contrast. The method presented in this paper is instead agnostic to the imaging modality and can 
be applied directly to the \textit{raw} data. This is because it models many 
features of the imaging process (bias field, gridding, etc.), in order not to
require any processing such as
skull-stripping, intensity normalisation, affine alignment or reslicing to a 
common grid.  These properties are important for a general tool that should 
work `out-of-the-box', given that imaging protocols are far from 
standardised -- 
restricting a method to a particular intensity profile considerably restricts 
its practical use. In addition, our method allows for a user to chose the resolution of the common space.
We validate our approach on a pairwise registration task, comparing it against 
state-of-the-art methods, on publicly available data. We achieve favourable 
results 
outperforming all other methods, within reasonable runtimes.

\begin{figure*}[t]
\centering
\scalebox{.8}{\begin{minipage}{.3\textwidth}
\hspace*{-1cm}  
\includegraphics{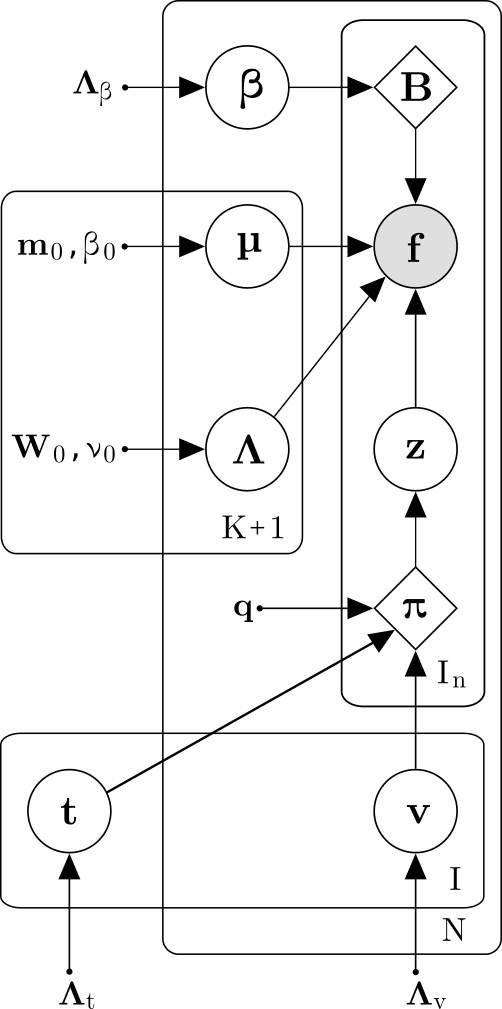}
\end{minipage}}
\scalebox{.8}{\begin{minipage}{.65\textwidth}
\begin{tabular}{rl}
\multicolumn{2}{c}{\textbf{Appearance} ($\mathcal{A}$)} \\
Intensity: & ${\bf f}_{ni} \sim \prod_k \mathcal{N}_C \left( 
{\bf B}_{ni}^{-1} \bm{\mu}_k, \left({\bf B}_{ni}\bm{\Lambda}_k{\bf B}_{ni}\right)^{-1}\right)^{z_{nik}}$\\
INU coefficients: & $\bm{\beta}_{nc} \sim \mathcal{N}_{M} 
\left( \bm{0}, 
\bm{\Lambda}_{\beta}\right)$\\
Intensity mean: & $\bm{\mu}_k \sim \mathcal{N}_C \left( {\bf m}_{0k}, 
(\beta_{0k} \bm{\Lambda}_k)^{-1}\right)$\\
Intensity precision: & $\bm{\Lambda}_k \sim \mathcal{W}_C \left( {\bf 
W}_{0k}, 
\nu_{0k}\right)$\\
INU function: & ${\bf B}_{ni} = \text{diag}({\bf b}_{ni}), \; {\bf 
b}_{ni} = \exp \left( \bm{\Upsilon}_i\bm{\beta}_{n} 
\right)$\\ 
Intensity parameters: & ${\bf m}_{0k}, 
\beta_{0k},{\bf W}_{0k}, \nu_{0k}$ \\
Hyper-parameters: & $\bm{\Lambda}_{\beta}\text{ (bending energy)}$\\[0.5em]
\multicolumn{2}{c}{\textbf{Shape} ($\mathcal{S}$)} \\
Tissue: & ${\bf z}_{ni} \sim \text{Cat}_{K+1}\left( \bm{\pi}_{ni} 
\right)$\\
Log-template: & ${\bf t}_k \sim \mathcal{N}_{I} 
\left( \bm{0}, 
\bm{\Lambda}_{\text{t}}\right)$\\
Velocity: & ${\bf v}_{n} \sim \mathcal{N}_{I} 
\left( \bm{0}, 
\bm{\Lambda}_{\text{v}}\right), \; \sum_n {\bf v}_n = 
\bm{0}$\\
Tissue prior: & $\bm{\pi}_{n} = \mathrm{softmax}\left({\bf t} \circ \bm{\psi}_{n} \right)$\\
Forward deformation: & $\bm{\psi}_{n} = \bm{\phi}({\bf v}_n) \circ {\bf M}_{\text{t}}{\bf R}({\bf q}_n){\bf M}_n^{-1}$\\
Rigid parameters: & ${\bf q}_n, \; \sum_n {\bf q}_n = 
\bm{0}$\\
Hyper-parameters: & $\bm{\Lambda}_{\text{v}}$, 
$\bm{\Lambda}_{\text{t}}$\text{ (combination of energies)}\\[0.5em]
\end{tabular}
\begin{tabular}{rlrl}
\multicolumn{4}{c}{\textbf{Number of}}\\
subjects: & $N$,&channels: & $C$\\
subject voxels: & $I_n$, & bias bases: & $M$\\
template voxels: & $I$ & template classes: & $K$
\end{tabular}\end{minipage}}
\caption{The joint probability distribution over $N$ images.
Random variables are in circles, observed are shaded, plates 
indicate replication, hyper-parameters have dots, diamonds indicate 
deterministic functions. The distributions in this figure are the 
Normal ($\mathcal{N}$), Wishart ($\mathcal{W}$) and Categorical 
($\text{Cat}$). Note that $K + 1$ mutually exclusive classes are modelled, but as the final class can be determined by the initial $K$, we do not represent it (improving runtime, memory usage and stability). The hyper-parameters ($\bm{\Lambda}_{\beta}$, 
$\bm{\Lambda}_{\text{v}}$, $\bm{\Lambda}_{\text{t}}$) encode a 
combination of absolute, membrane and bending energies. 
$\bm{\Lambda}_{\text{v}}$ further penalises linear-elasticity. The sum 
of the shape parameters (${\bf v}_n$, ${\bf q}_n$) are constrained to 
zero, to ensure that the template remains in the average position 
\cite{beg2006computing}.}
\label{fig:gm}
\end{figure*}

\section{Methods}

\textbf{Generative Model.} In this work, computing the nonlinearly 
aligned images is actually a 
by-product of doing inference on a joint probability distribution. This 
generative model consists of multiple random variables, modelling 
various properties of the observed data. It is defined by the following 
distribution:
\begin{align}
p(\mathcal{F},\mathcal{A},\mathcal{S}) = 
p(\mathcal{F} \mid \mathcal{A}, \mathcal{S})~
p(\mathcal{A}, \mathcal{S}),
\label{eq:model}
\end{align}
where $\mathcal{F} = \{{\bf F}_n\}_{n=1}^N$, ${\bf F}_n \in 
\mathbb{R}^{I_n \times C}$ are the $N$ observed images (\emph{e.g.}, MRI scans), each with $I_n$ 
voxels 
and $C$ channels (\emph{e.g.}, MR contrasts). The two sets $\mathcal{A}$ and $\mathcal{S}$ 
contain the appearance and shape variables, respectively. The 
distribution in \eqref{eq:model} is unwrapped in detail in Fig. 
\ref{fig:gm}, showing its graphical model and constituent parts.
The inversion of the model in \eqref{eq:model} is performed using
a variational 
expectation-maximisation (VEM) algorithm. In this algorithm, each 
parameter (or its probability distribution, in the case of the mixture 
parameters) is updated whilst holding all others fixed, in an 
alternating 
manner \cite{bishop2006pattern}. The individual update equations are 
obtained from 
the evidence lower bound (ELBO):
\begin{align}
\mathcal{L} = \sum_{\mathcal{A},\mathcal{S}} q(\mathcal{A},\mathcal{S}) 
\ln \left[ 
\frac{p(\mathcal{F},\mathcal{A},\mathcal{S})}{q(\mathcal{A},\mathcal{S})}
 \right],
\label{eg:elbo}
\end{align}
where the variational distribution is assumed to factorise as 
$q(\mathcal{A},\mathcal{S}) = q(\mathcal{A})q(\mathcal{S})$. The 
appearance updates have been published in previous work: the 
inference of the intensity parameters in 
\cite{blaiotta2018generative}; the mode 
estimates of the intensity non-uniformity (INU) parameters in 
\cite{ashburner2005unified}.  

The contribution of this paper is to unify the shape and appearance parts as \eqref{eq:model}, providing a flexible and unsupervised image registration framework. In particular, this framework relies on: parameterising the shape model using a combined rigid and diffeomorphic registration in the space of the template, introduction of a 
multi-scale optimisation method, and a 
novel way of computing a Hessian of 
the categorical data term. These will next be explained in more detail. \\

\noindent\textbf{Spatial Transformation Model.}
For maximum generalisability, the model should handle image data defined on 
arbitrary lattices with arbitrary orientations (\emph{i.e.}, any 
well formatted NIfTI file).
The forward deformation $\vec{\psi}_n$, warping the template 
to subject space, is the composition of a diffeomorphic 
transform ${\bm \phi}_n$, defined over the template field of view, 
and a rigid transform ${\bf R}_n$, defined in world space. The 
template (${\bf M}_{\text{t}}$) and subject (${\bf M}_n$) orientation matrices describe the mapping from voxel to world 
space. Therefore, 
${\bm \psi}_n = {\bm \phi}_n \circ {\bf M}_{\text{t}} \circ {\bf R}_n \circ {\bf M}_n^{-1}$. 
The diffeomorphism is encoded by the initial velocity of the 
template `particles' \cite{ashburner2019algorithm}, and
recovered by geodesic shooting \cite{miller2006geodesic}: ${\bm \phi}_n 
= 
\mathrm{shoot}\left({\bf v}_n\right)$. ${\bf 
R}_n$ is encoded by its projection $\mathbf{q}_n$
on the tangent space of rigid transformation matrices, and recovered by 
matrix exponentiation \cite{woods2003characterizing}. ${\bf R}_n$ could have included scales and shears, but keeping it rigid allows us to capture these deformations in the velocities.\\

\noindent\textbf{Multi-Scale Optimisation.}
Registration is a non-convex problem and is therefore highly sensitive to 
local minima. Multi-scale optimisation techniques can be used to circumvent this problem \cite{christensen1997volumetric,zollei2005efficient,krebs2019learning}. The proposed 
approach implements such a multi-scale method to help with 
several difficulties: local minima (especially in the rigid parameter space), 
slow VEM convergence, and slow runtime. The way we parameterise the spatial transformation model is what enables our multi-scale approach. If we drop all terms that do not depend on the
template, velocities or rigid parameters, the ELBO in \eqref{eg:elbo} reduces to:
\begin{align}
\mathcal{L} \cequal \sum_n \Big\{ 
\ln \mathrm{Cat}\left(\tilde{\mathbf{z}}_n \mid \mathrm{softmax}\left({\bf t} \circ \bm{\psi}_{n} \right)\right)
+ \ln p(\vec{v}_n)
\Big\}
+ \ln p({\bf t})
~,
\end{align}
where $\tilde{\mathbf{z}}_n$ denotes the latent class posterior probabilities 
(responsibilities). The two prior terms originate from the realm of 
PDEs, where they take the form of integrals of continuous functions. When 
discretised, these integrals can be interpreted as negative logs of 
multivariate Normal distributions (up to a constant):
\begin{align}
\frac{\lambda}{2} \int_{\Omega} \langle f(\mathbf{x}), (\Lambda f)(\mathbf{x}) \rangle d\mathbf{x}\;
\xrightarrow{\text{discretise}}\;
\frac{\lambda}{2} \left(\mathbf{f}^{\mathrm{T}}\bm{\Lambda}\mathbf{f}\right) 
~\Delta_x.
\end{align}
Here, $\mathbf{f}^{\mathrm{T}}\bm{\Lambda}\mathbf{f}$ computes the 
sum-of-squares of the (discrete) image gradients and $\Delta_x$ is the volume 
of one discrete element. Usually, $\Delta_x$ would simply be merged into the 
regularisation factor $\lambda$. In a multi-scale setting, it must be correctly 
set at each scale. In practice, the template and velocities are first defined
over a very coarse grid, and the VEM scheme is applied with a suitable scaling.
At convergence, they are trilinearly interpolated to a finer grid, and the scaling parameter 
is changed accordingly for a new iteration of VEM.\\


\noindent\textbf{B\"ohning Bound.}
We use a Newton-Raphson algorithm to find mode estimates of the variables $\mathbf{t}$, $\mathbf{v}_n$ and $\mathbf{q}_n$, with high convergence rates. This requires the gradient and Hessian of the categorical data term.
If the gradient and Hessian with respect 
to $\mathbf{t}_n = {\bf t} \circ \bm{\psi}_{n}$ are known, then those with 
respect to the variables of interest $\mathbf{t}$, $\mathbf{v}_n$
and $\mathbf{q}_n$ can be obtained by application of the chain rule 
(with Fisher's scoring \cite{ashburner2019algorithm}). 
However, the true Hessian is not well-behaved 
and the Newton-Raphson iterates may overshoot. Therefore, some 
precautions must be taken such as ensuring monotonicity using 
a backtracking line search \cite{ashburner2009computing}.
Here, we make use of B\"ohning's approximation \cite{bohning1992multinomial} 
to bound the ELBO and improve the  stability of the update steps, without the 
need for line search. This approximation was introduced in the context of 
multinomial logistic regression, which relies on a similar objective function. 
Because this approximation allows the true objective function to be bounded, it
ensures the sequence of Newton-Raphson steps to be monotically improving. 
However, this bound is not quite tight, leading to slower convergence rates. 
In this work, we therefore use a weighted average of B\"ohning's
approximation and the true Hessian that leads to both fast and stable convergence; 
\emph{e.g.}, the template Hessian becomes:
\begin{align}
\frac{\partial^2\mathcal{L}}{\partial t_{nik} \partial t_{nil}} \approx w \underbrace{\pi_{nik} \left( \delta_k^l - \pi_{nil} \right)}_{\text{True Hessian}} {}+{} (1 - w)  \underbrace{\frac{1}{2}\left( \delta_k^l - \frac{1}{K}\right)}_{\text{B\"ohning bound}}, \quad w \in [0,1].
\end{align}

\section{Validation}

\noindent \textbf{Experiments.} 
Brain scans where regions-of-interests have been
manually labelled by human experts can be used to assess the accuracy of a registration method. By warping the label images from one subject onto another, overlap scores can be computed, without the need to resample the groundtruth annotations. The labels parcelate the brain into small regions, identifying the same anatomical structures between subjects. As the labels are independent from the signal used to compute the deformations, they are well suited to be used for validation. Such a validation was done in a seminal paper \cite{klein2009evaluation}, where 14 methods were compared at nonlinearly registering pairs of MR brain scans. Two datasets used in \cite{klein2009evaluation} were\footnote{\url{nitrc.org/projects/ibsr}, \url{resource.loni.usc.edu/resources}}:
\begin{itemize}

\item
\textbf{LPBA40}:
T1-weighted (T1w) MRIs of 40 subjects with cortical and 
subcortical labels, of which 56 were used in the validation in 
\cite{klein2009evaluation}. The two top-performing methods, from 
$N=1,560$ pairwise registrations, were ART's 3dwarper 
\cite{ardekani2005quantitative} and ANTs' SyN 
\cite{avants2008symmetric}. The MRIs have been processed by 
skull-stripping, non-uniformity correction, and rigid reslicing to a 
common space.

\item
\textbf{IBSR18}:
T1w MRIs of 18 subjects with cortical
labels, where 96 of the labelled regions were
used in the validation in \cite{klein2009evaluation}. The two top-performing methods, from
$N=306$ pairwise registrations, were SPM's Dartel
\cite{ashburner2007fast} and ANTs' SyN \cite{avants2008symmetric}. The
MRIs have non-isotropic voxels and are unprocessed; IBSR18 are 
therefore more challenging to register than LPBA40.

\end{itemize}

We now compare our method, denoted MultiBrain (MB), with the 
top-performing methods in \cite{klein2009evaluation}, on IBSR18 and 
LPBA40. The same overlap metric is used: the volume over which the 
deformed source labels match the target labels, divided by the total 
volume of the target labels (\emph{i.e.}, the true positive rate 
(TPR)). 
Two additional registration methods are included: one state-of-the-art group-wise model, SPM's Shoot
\cite{ashburner2011diffeomorphic}; and one state-of-the-art deep 
learning model, the CVPR version of 
VoxelMorph\footnote{\url{github.com/voxelmorph/voxelmorph}} (VXM) 
\cite{balakrishnan2019voxelmorph}. 
Pairwise registrations between all subjects (in both directions) are 
computed using MB, SPM's Shoot and VXM. For MB, we (i) learned the 
optimal average from each dataset (MB-GW), and (ii) learned the optimal 
average from a held-out training set (MB-L); as described in Fig. 
\ref{fig:RegMethods}b. These two tasks are similar to either finding 
the optimal template for a specific neuroimaging group study (MB-GW) or 
using a predefined common space for the same task (MB-L). Shoot's 
registration process resembles MB-GW, whilst VXM resembles MB-L. MB-L 
was trained on $N=277$ held-out T1w MRIs from five different 
datasets: three publicly 
available\footnote{\url{brain-development.org}, 
\url{mrbrains18.isi.uu.nl}, \url{my.vanderbilt.edu/masi}}:
 IXI ($N_1=200$), 
MICCAI2012 ($N_2=35$) and MRBrainS18 ($N_3=7$); and two hospital 
curated ($N_4=19$, $N_5=16$). Training took two days on a modern 
workstation.

%
%
%

\begin{figure*}[t]
\centering
\includegraphics[width=\textwidth]{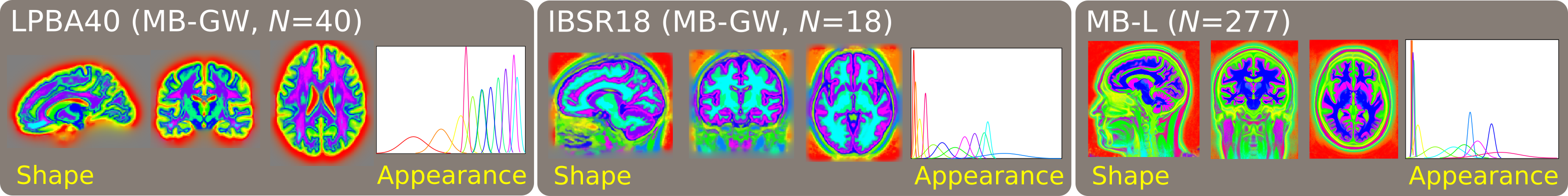}
\caption{Learned shape and appearance priors from fitting MB-GW to LPBA40 (left) and IBSR18 (middle); and MB-L to
a training dataset (right). Colours
correspond to clusters found, unsupervised, by fitting the model.
Appearance densities show the expectations of the Gaussians drawn from
the Gauss-Wishart priors (using $3$ $\sigma$).}
\label{fig:Models}
\end{figure*}

The shape and appearance models that were learned when fitting MB are shown in Fig.
\ref{fig:Models}. $K=11$ classes were used, 1 mm
isotropic template voxels and the priors were initialised as uninformative. The initial template and velocity dimensions were set to 8 mm cube. Energy hyper-parameters were chosen as
$\lambda_{\beta}=1\text{e}5$, $\lambda_{\text{v}}=\{2\text{e-}4, 0, 0.4, 0.1,
0.4\}$ (absolute, membrane, bending, linear elasticity) and $\lambda_{\text{t}}=\{1\text{e-}2, 0.5, 0\}$ (absolute, membrane, bending). The weighting was set to $w=0.8$. The algorithm was run for a predefined number of iterations.\\

\begin{figure*}[t]                                                 
\centering
\includegraphics[width=1\textwidth]{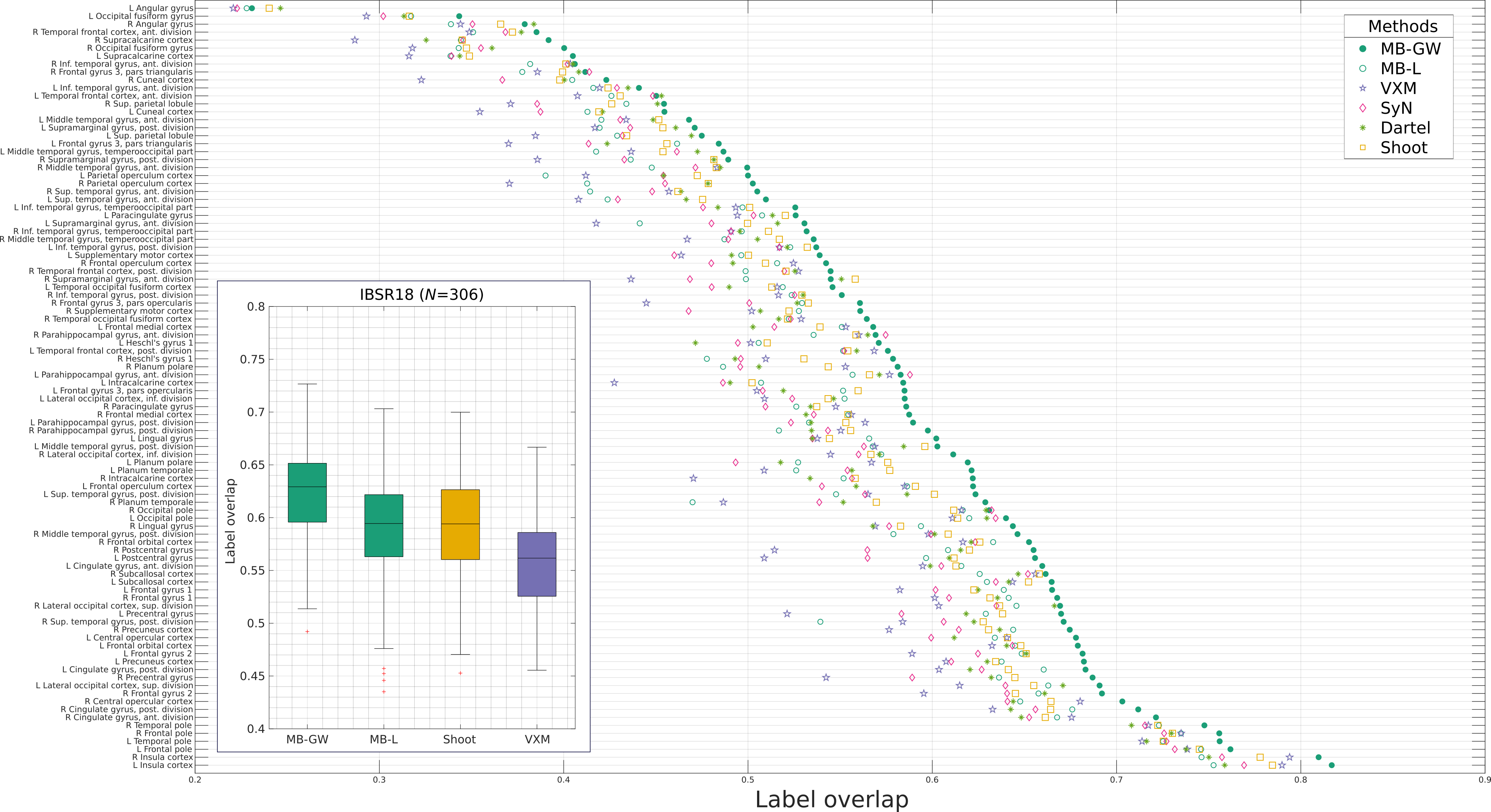}
\caption{Results from the validation on the IBSR18 dataset. 
The nonlinear registration methods include MB-GW/L, SPM's Shoot, VXM and 
the two top algorithms evaluated in
\cite{klein2009evaluation}. Shown are the average label overlaps and total overlaps (the
boxplot). The results in the boxplot may be compared directly with
the methods of Fig. 5 in \cite{klein2009evaluation}.}
\label{fig:Labels-IBSR18}
\end{figure*}

\noindent \textbf{Results.}
The label overlap scores on IBSR18 are shown in Fig. 
\ref{fig:Labels-IBSR18}. The figure shows, close to, 
unanimous better overlap for MB, compared to the other algorithms. 
Result plots for LPBA40 are given in the supplementary materials, 
as well as samples of the best and worst registrations for MB and VXM. 
On both IBSR18 and LPBA40, MB performs favourably. For IBSR18, the mean and median 
overlaps were 0.62 and 0.63 respectively for MB-GW, and both 0.59 for MB-L. Mean and median overlaps were 
both 0.59 for SPM's Shoot and both 0.56 for VXM. The greatest median 
overlap reported in \cite{klein2009evaluation} was about 0.55, whereas 
the overlap from affine registration was 0.40 
\cite{jenkinson2002improved}. For LPBA40, the mean and 
median overlaps were both 0.76 for MB-GW and both 0.75 for MB-L. 
Mean and median overlaps for SPM's Shoot approach 
were both 0.75, and both 0.74 for VXM. The highest median 
overlap reported in \cite{klein2009evaluation} was 0.73, and that from 
affine registration was 0.60 \cite{jenkinson2002improved}. 
Using the affine registrations as baseline, the results showed 6\% to 
17\% greater accuracy improvements when compared to those achieved for 
the second most accurate nonlinear registration algorithm
evaluated\footnote{$(\text{TPR}_{\mathrm{MB}} - 
\text{TPR}_{\mathrm{Shoot}})/(\text{TPR}_{\mathrm{MB}} - 
\text{TPR}_{\mathrm{Affine}}) \times 
100\%$}. 
Computing one forward deformation took about 15 minutes for MB-L and 30 for MB-GW
(on a modern workstation, running on the CPU).\\

\noindent \textbf{Discussion.} 
MB-GW does better than MB-L, this
was expected as the average obtained by groupwise fitting directly on
the population of interest should be more optimal than one learned from
a held-out dataset, on a limited number of subjects (\emph{e.g.}, the
averages for the individual populations in Fig. \ref{fig:RegMethods} are
more optimal than the global). Still, MB-L learned on
only 277 subject does as well as, or better than, Shoot (a
state-of-the-art groupwise approach). This is an exciting result that
allows for groupwise accuracy spatial normalisation on small number of
subjects, and to a standard common space (instead of a
population-specific). With a larger and more diverse training
population, accuracies are expected to improve further. One may claim
that a group-wise registration scheme has unfair advantage over pairwise
methods. However, as a common aim often is to spatially normalise - with
the objective of making comparisons among a population of scans, it
would be reasonable to aim for as much accuracy as possible for this
task. The purely data-driven VXM approach does better than the methods
evaluated in \cite{klein2009evaluation}. VXM was trained on close
to 4,000 diverse T1w MRIs. A larger training dataset could boost its
performance. The processing that was applied to the VXM input data was
done using SPM \cite{malone2015accurate}, whilst its training data was
processed using FreeSurfer. Having used the same software could have
improved its results; however, being reliant on a specific processing
pipeline is inherently a weakness of any method. Furthermore, the VXM
model uses a cross-correlation loss function that should be resilient to
intensity variations in the T1w scans. Finally, the contrasts and 
fields of view in the 
T1w scans were slightly different from each other in the training and
testing data, due to variability in field strength and scanner settings.
This could have impacted the accuracy of MB-L and VXM.

\section{Conclusion}

This paper introduced an unsupervised learning algorithm for
nonlinear image registration, which can be applied to unprocessed 
medical imaging data. A validation on two publicly available datasets 
showed state-of-the art results on registering MRI brain scans. 
The unsupervised, non-organ specific nature of the algorithm makes 
it applicable to not only brain data, but also other types of medical 
images. This could allow for transferring methods widely used in 
neuroimaging to other types of organs, \emph{e.g.}, the liver 
\cite{ridgway2018voxel}. The runtime of the algorithm is not on par 
with a GPU implementation of a deep learning model, but still allows 
for processing of a 3D brain scan in an acceptable time. The runtime should
furthermore improve, drastically, by an implementation on the GPU. 
The proposed model could also be used for image segmentation 
\cite{ashburner2005unified} and translation \cite{brudfors2019empirical}, or modified to use labelled data, in a semi-supervised manner \cite{blaiotta2018generative}. Finally, the multi-modal 
ability of the model would be an interesting avenue of further research. 



\subsubsection*{Acknowledgements:} MB was funded by the EPSRC-funded
UCL Centre for Doctoral Training in Medical Imaging (EP/L016478/1) and
the Department of Health’s NIHR-funded Biomedical Research Centre at
University College London Hospitals. YB was funded
by the MRC and Spinal Research Charity through the ERA-NET Neuron joint
call (MR/R000050/1). MB and JA were funded by the EU
Human Brain Project's Grant Agreement No 785907 (SGA2). GF and the Wellcome Centre for Human Neuroimaging is supported by core funding from the Wellcome (203147/Z/16/Z).

\bibliography{bibliography}%
\bibliographystyle{ieeetr}%

\clearpage
\section*{Supplementary Materials}

\begin{figure*}[h!]                                            
	\centering
	\includegraphics[width=1\textwidth]{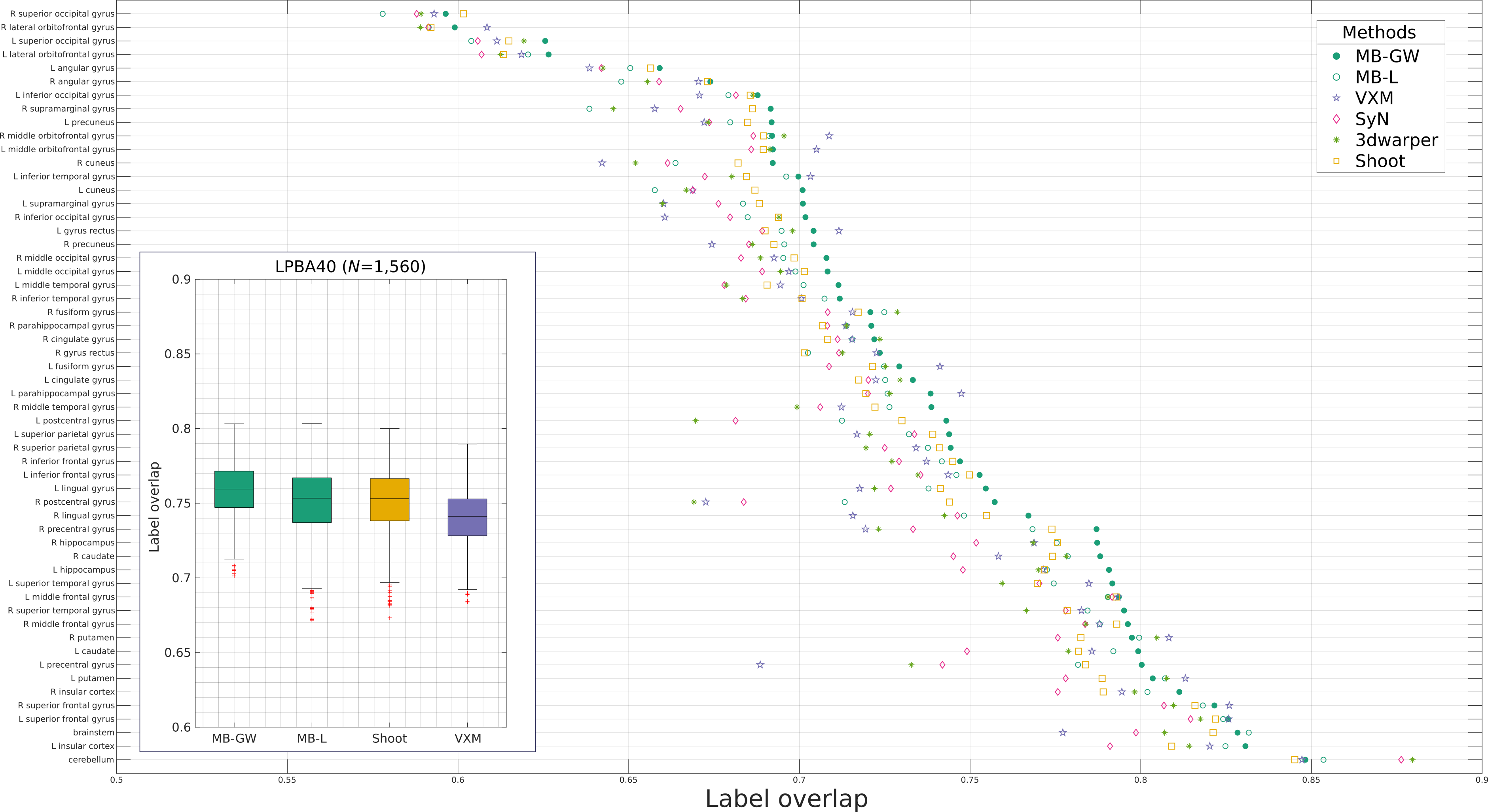}
	\caption{Results from the validation on the LPBA40 dataset. 
		The nonlinear registration methods include MB-GW/L, SPM's Shoot, VXM and 
		the two top algorithms evaluated in
		\cite{klein2009evaluation}. Shown are the average label overlaps and total overlaps (the
		boxplot). The results in the boxplot may be compared directly with
		the methods of Fig. 5 in \cite{klein2009evaluation}. On each box, the central mark is the median, the edges of the
		box are the 25th and 75th percentiles, the whiskers extend to the most
		extreme data-points not considered outliers. Any outliers are plotted
		individually.}
	\label{fig:Labels}
\end{figure*}

\begin{figure*}[h!]
	\centering
	\includegraphics[width=1\textwidth]{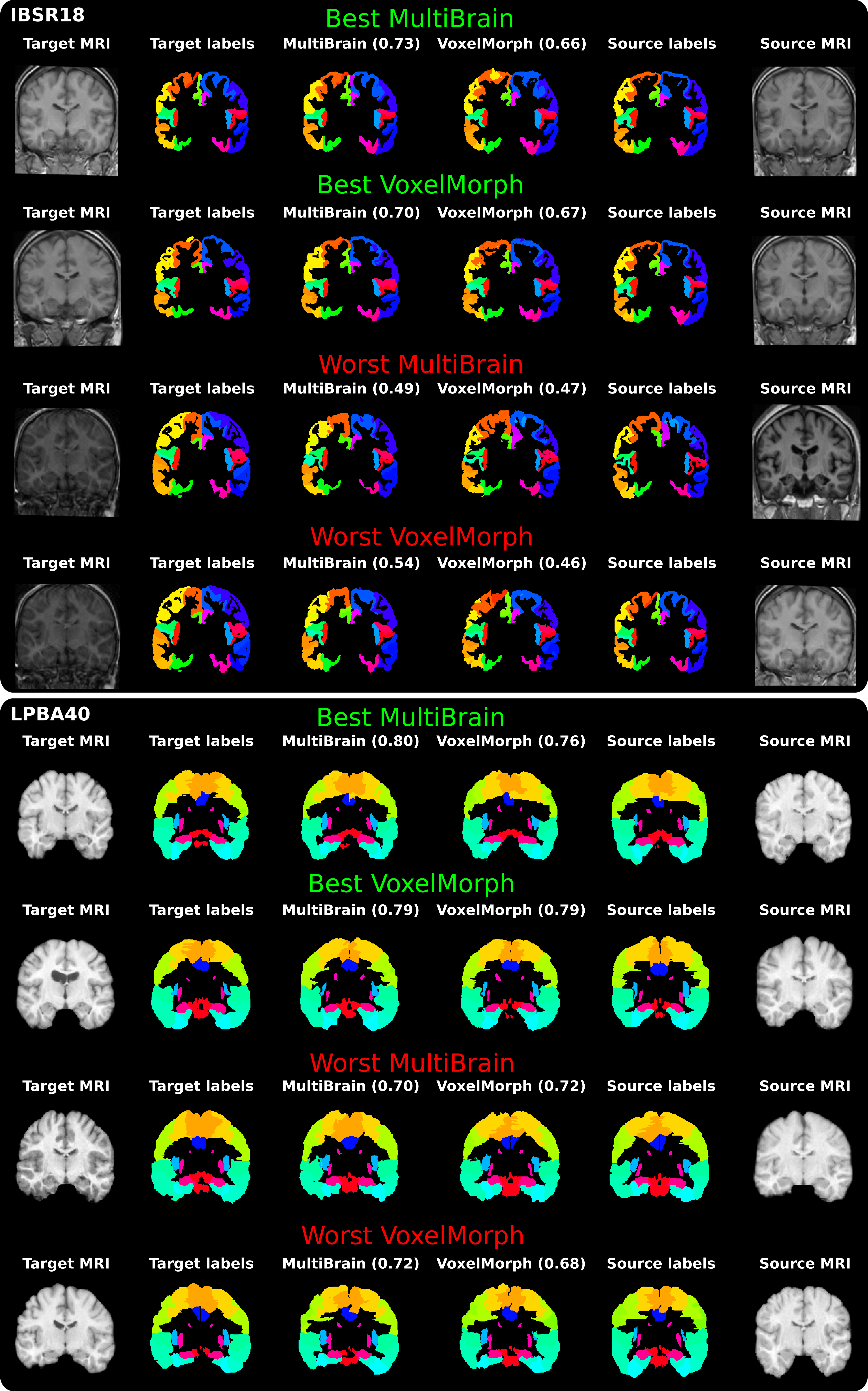}
	\caption{Registrations with best and worst total overlap scores, for
		MB-L and VXM, on the IBSR18 (top) and LPBA40
		(bottom) datasets. Shown are: target and source MRIs+labels; and source
		labels warped to target labels, for both methods (overlaps in
		parenthesis).}
	\label{fig:BestWorst}
\end{figure*}

\end{document}